\documentclass[]{sig-alternate-2013}
\usepackage[square,sort,comma,numbers]{natbib}
\usepackage[utf8]{inputenc}
\usepackage{amsmath}
\usepackage{color}
\usepackage{natbib}
\usepackage{graphicx}
\usepackage{multirow}
\usepackage{subcaption}
\usepackage{booktabs}     
\usepackage{algorithm}
\usepackage{algorithmic}
\usepackage{etoolbox}
\usepackage{url}
\patchcmd{\maketitle}{\@copyrightspace}{}{}{}
\makeatother
\newcommand{\algrule}[1][.2pt]{\par\vskip.5\baselineskip\hrule height #1\par\vskip.5\baselineskip}

\DeclareMathOperator*{\argmin}{arg\,min}
\newcommand{\wi}{\mathbf{w}_{i}}
\newcommand{\hj}{\mathbf{h}_{j}}
\newcommand{\wit}{\mathbf{w}_{i}^{T}}
\newcommand{\hjt}{\mathbf{h}_{j}^{T}}
\newcommand{\wihj}{\mathbf{w}_{i}\mathbf{h}_{j}^{T}}
\newcommand{\R}{\mathcal{R}}
\newcommand{\Ri}{\mathcal{R}_{i\bullet}}
\newcommand{\Rj}{\mathcal{R}_{\bullet j}}
\newcommand{\spara}[1]{\smallskip\noindent{\bf #1}}
\newcommand{\dataset}[1]{\textsf{#1}}
\newcommand{\codeurl}{{\small \url{https://github.com/rdevooght/MF-with-prior-and-updates}}}
\newcommand{\SL}{squared loss}
\newcommand{\AL}{absolute loss}
\newcommand{\GKL}{generalized Kullback-Liebler divergence}
\newcommand{\VW}{VW}

\newcommand{\superscript}[1]{\ensuremath{^{\textrm{#1}}}}
\def\sharedaffiliation{\end{tabular}\newline\begin{tabular}{c}}

\newenvironment {squishlist}
{\begin{list}{$\bullet$}
  { \setlength{\itemsep}{1pt}
     \setlength{\parsep}{1pt}
     \setlength{\topsep}{1pt}
     \setlength{\partopsep}{1pt}
     \setlength{\leftmargin}{1.5em}
     \setlength{\labelwidth}{1em}
     \setlength{\labelsep}{0.5em} } }
{\end{list}}

\title{Dynamic Matrix Factorization with Priors\\on Unknown Values}

\numberofauthors{1}
\author{
\alignauthor Robin Devooght\titlenote{This work was carried out while the author was an intern at Yahoo Labs, Barcelona.} \ \ Nicolas Kourtellis\titlenote{This work was carried out while the author was with Yahoo Labs, Barcelona.}  \ \ Amin Mantrach\superscript{**} \\
\sharedaffiliation
{robin.devooght@ulb.ac.be, nicolas.kourtellis@telefonica.com, amantrac@yahoo-inc.com}\\
  \begin{tabular}{cccc}
	\affaddr{\superscript{*}IRIDIA, ULB}	\;\;\;\;& \affaddr{\superscript{\dag}Telefonica Research }	\;\;\;\;&	\affaddr{\superscript{**}Yahoo Labs} \\
	\affaddr{1050 Brussels, Belgium}		\;\;\;\;&	\affaddr{08019 Barcelona, Spain} \;\;\;\;&	\affaddr{08018 Barcelona, Spain} \\
  \end{tabular}
}

\begin{document}

\maketitle

\begin{abstract}
Advanced and effective collaborative filtering methods based on explicit feedback assume that unknown ratings do not follow the same model as the observed ones (\emph{not missing at random}). 
In this work,  we build on this assumption,  and introduce a novel dynamic matrix factorization framework that allows to set  an explicit prior on unknown values.
When new ratings, users, or items enter the system, we can update the factorization in time independent of the size of data (number of users, items and ratings).
Hence, we can quickly recommend items even to very recent users.
We test our methods on three large datasets, including two very sparse ones, in static and dynamic conditions. 
In each case, we outrank state-of-the-art matrix factorization methods that do not use a prior on unknown ratings.
\end{abstract}

\section{Introduction}\label{sec:intro}
Personalizing the user experience is a continuous growing challenge for various digital applications. 
This is of particular importance when  recommending releases on the Netflix platform, when digesting latest Yahoo news, or for  helping users to find their next musical obsession. 

Among the different approaches towards personalization, matrix factorization ranges among the most popular ones \cite{koren2009matrix, zhou2008large}. 
In this line of work, data is represented in the form of a user-item matrix, encoding user-item interactions in the form of binary or real values. 
Matrix factorization aims at decomposing a matrix into latent representations designed to accurately reconstruct observed interaction values. 
Most interestingly, these latent features are also used to predict missing (or unknown) ratings (i.e. if item $j$ is exposed to user $i$, what would be his rating). 
However, by trying to predict the unknown ratings based on a model trained on the observed ratings, the recommender systems implicitly assume that the distribution of the observed ratings is representative of the distribution of the unknown ones. 
This is called the \emph{missing at random} assumption \cite{little2002statistical}, and it is probably a wrong asumption in most real-world applications. 
In the case of a movie recommender system, for example, users rate movies that they have seen, and their choices are biased by their interests.

In this work, building on the \emph{not missing at random assumption} ~\cite{missing_at_random, steck2010training} we make the hypothesis that it is more likely for an unknown item to be weakly rated, this due to the huge amounts of existing items coupled to the limited number of items a user may be interested in. 
This translates into a strong prior suggesting that unknown ratings should be reconstructed from latent features as small values (i.e. close to 0).
While this assumption may be wrong for specific cases, such constraints act as a good regularizer that helps in significantly improving the recommendations. 

Our work is not the first to propose new interpretation of the missing data in a matrix factorization framework \cite{implicit_feedback_confidence, ALS_unknown_weighting, weighting_sampling,steck2010training}. 
However, to the best of our knowledge, we are the first to propose an \emph{online learning} mechanism that sets an explicit prior on unknown values and this, without any significant additional cost.
We introduce a method to update our model each time a new rating is observed with a time complexity independent of the size of the data (i.e. the total number of users, items, and ratings). 
This fast update mechanism allows keeping the model up to date when a flow of new users, items and ratings enters the system. 

The contributions of this work are as follows:
\begin{squishlist}

\item We extend the \SL, the \AL\ and the \GKL\ to take into account an explicit prior on unknown values. 

\item For each loss function, we derive an efficient \emph{online learning algorithm} to update the parameters of the model with a complexity independent of the data size.

\item We validate the hypothesis that applying an explicit prior on missing ratings improves the recommendations in a static and in a dynamic setting on three public datasets.

\item Our methods are easy to implement and we provide an open-source implementation of the \SL~and \AL. 
\end{squishlist}

The rest of this paper is organized as follows.
Section~\ref{sec:recommendation-problem} summarizes the recommendation problem and Section~\ref{sec:missing-data} formulates how to apply priors on unknown values in the context of recommendation.
Section~\ref{sec:loss-funtcions} extends three loss functions and shows how they can be optimized in a static and dynamic fashion.
Section~\ref{sec:experiments} presents our experimental results and Section~\ref{sec:related-work} discusses works related to our study.
Section~\ref{sec:conclusion} concludes this paper.
\section{The recommendation problem}\label{sec:recommendation-problem}

Before addressing the challenge of interpreting missing data, let us state the standard recommendation problem. 

We have at our disposal  $m$ items rated by $n$ different users, where the rating given by the $i^{\text{th}}$ user to the $j^{\text{th}}$ item is denoted by $r_{ij}$. 
In many real applications, these ratings take an integer value between 1 and 5. 
In this work, we assume that ratings are positive and that an item rated by user $i$ with a high numerical value is preferred by this user over items she ranked with lower numerical values. 
We denote by $\R$ the set of all known ratings, and by $\Ri$ and $\Rj$ the set of known ratings of user $i$ and item $j$, respectively.
If $r_{ij} \notin \mathcal{R}$ we say that the rating is \emph{unknown}.

For a while, the objective of recommender systems has been to predict the \emph{value} of unknown ratings \cite{koren2009matrix}. 
It is now widely accepted that a more practical goal is to correctly \emph{rank} the unknown ratings for each user, while the actual value of the rating is of little interest \cite{balakrishnan2012collaborative, cremonesi2010performance, lee2014local, ALS_unknown_weighting}.
This has led to a change in the way methods are evaluated (in terms of ranking metrics such as NDCG, AUC or MAP, instead of rating prediction metrics as measured by RMSE).
We embrace that shift towards \emph{ranking}, and the purpose of adding a prior on the unknown ratings is not to improve matrix factorization techniques in terms of RMSE, but in terms of ranking metrics.

Matrix factorization methods produce for each user and each item a vector of $k$ ($<< n$ and $m$) real values that we call \emph{latent features}.
We denote by ${w}_{i}$ the row vector containing the $k$ features of the $i^{\text{th}}$ user, and $h_{j}$ the row vector, composed of $k$ features, associated to the $j^{\text{th}}$ item.
Also, we denote by $\mathbf{W}$ the $n \times k$ matrix whose $i^{\text{th}}$ row is $\wi$, and $\mathbf{H}$ as the $k \times m$ matrix whose $j^{\text{th}}$ column is $\hjt$.
Matrix factorization is presented as an optimization problem, whose general form is:

\begin{align}
\argmin_{\mathbf{W}, \mathbf{H}} \sum_{i,j|r_{ij} \in \mathcal{R}} E\left( r_{ij}, \wihj \right) + R(\mathbf{W}, \mathbf{H})
\label{eq:classic_problem}
\end{align}
where $R$ is a regularization term (often $L_{1}$ or $L_{2}$ norms), and $E$ measures the error that the latent model makes on the observed ratings.
Most often, $E$ is the squared error.

Using a matrix factorization approach for predicting unknown ratings relies on the hypothesis that a model accurately predicting observed rating generalizes well to unknown ratings.
In the following section, we argue that the former hypothesis is easily challenged.
\section{Interpreting missing data}\label{sec:missing-data}

LaunchCast is Yahoo's former music service, where users could, among other things, rate songs.
In a survey of 2006, users were asked to rate randomly selected songs \cite{missing_at_random}.
The distribution of ratings of random songs was then compared to the distribution of voluntary ratings.
The experiment concluded that  the distribution of the ratings for random songs was strongly dominated by low ratings, while the voluntary ratings had a distribution close to uniform  \cite{missing_at_random}.

Intuitively, a simple process could explain the results: users chose to rate songs they listen to, and listen to music they expect to like, while avoiding genres they dislike.
Therefore, most of the songs that would get a bad rating are not voluntary rated by the users.
Since people rarely listen to random songs, or rarely watch random movies, we should expect to observe in many areas a difference between the distribution of ratings for random items and the corresponding  distribution for the items selected by the users. 
This observation has a direct impact on the presumed capacity of matrix factorization to generalize a model based on observed ratings to unknown ratings.

Building on the \emph{not missing at random assumption} ~\cite{missing_at_random,steck2010training},  we propose to incorporate in the optimization problem stated in Equation \ref{eq:classic_problem} a prior about the unknown ratings, in order to limit the bias caused by learning on observed ratings:

\begin{align}
\begin{split}
\argmin_{\mathbf{W}, \mathbf{H}} & \sum_{i,j|r_{ij} \in \mathcal{R}} E\left( r_{ij}, \wihj \right) \\ & + \alpha \sum_{i,j|r_{ij} \notin \mathcal{R}} E\left( \hat{r}_{0}, \wihj \right) + R(\mathbf{W}, \mathbf{H})
\end{split}
\label{eq:prior_problem}
\end{align}

The objective function (Equation \ref{eq:prior_problem}) has now two parts (besides the regularization): the first part fits the model to the observed ratings, and the second part drives the model toward a prior estimate $\hat{r}_{0}$ on the unknown ratings. 
In absence of further knowledge about a specific dataset, we suggest to use $\hat{r}_{0} = 0$, the worst rating, as a prior estimate. 
The coefficient $\alpha$ allows to balance the influence of the unknown ratings, and the original formulation is obtained with $\alpha = 0$. 
We expect $\alpha$ to be small to deal with the problem of class imbalance. 
Indeed, in real-life applications the number of known ratings $|\mathcal{R}|$ is very small in comparison to the number of unknown ratings ($nm - |\mathcal{R}|$), and if $\alpha$ is close to 1, or larger, the second term of the objective function will completely dominate the other parts and drive all the users' and items' features to zero. 
It is therefore important to find a right balance between the influence of the few known ratings and of the many unknown ones. 

In order to have a more intuitive feeling of the influence of both parts of the objective function we introduce $\rho = \alpha (nm - |\mathcal{R}|) / |\mathcal{R}|$, which can be interpreted as an influence ratio between unknown and known ratings. 
If $\rho = 0$, the unknown ratings are ignored, if $\rho = 1$, both the known ratings and the unknown ratings have the same global influence on the objective function, if $\rho = 2$, the unknown ratings are twice as important as the known ratings, etc.

A more involved model could assume an adaptive $\rho$ per user or item, which could lead to additional, albeit small, gains.
However, this implies more parameters to tune, more cumbersome equations to explain and an involved process to prove that the complexity of the method remains the same.
Due to limited space, instead, we provide a general demonstration of the method and leave the adaptive model for future work.
\section{Loss functions}\label{sec:loss-funtcions}

An obvious difficulty raised by the new optimization problem introduced earlier is the apparent increase in complexity. 
The naive complexity of evaluating this objective function is $O(nmk)$, while it is $O(|\R|k)$ for classical matrix factorization approaches (Equation \ref{eq:classic_problem}).
In this section, we demonstrate how it is possible to use our new model without the naive additional cost, and present a way to perform fast updates to incorporate new ratings in the model.

To this end, we show the applicability of our method when $E$ is the \SL~in Section~\ref{sec:square-loss} and the \AL~in Section~\ref{sec:absolute-loss}.
For the sake of demonstration, we also discuss its applicability on the \GKL~in Section~\ref{sec:kullback-leibler}.
Finally, in Section~\ref{sec:static-dynamic} we outline how the method can be enforced in a static setting, and a dynamic setting with continuous updates of new ratings, items and users.

\subsection{Squared Loss}\label{sec:square-loss}

By  considering $E$ as the \SL, and $R$ as the $L_{1}$ regularization, the optimization problem becomes:

\begin{align}
\begin{split}
\argmin_{\mathbf{W}, \mathbf{H}} & \sum_{i,j|r_{ij} \in \R} \left( r_{ij} - \wihj \right) ^{2} \\ & + \alpha \sum_{i,j|r_{ij} \notin \R} \left( \wihj \right) ^{2} \\ & + \lambda \left( \sum_{i = 1}^{n} || \wi || _{1} + \sum_{j = 1}^{m} || \hj || _{1} \right)
\end{split}
\label{eq:opt_problem}
\end{align}

For the sake of simplicity, let us forget about the regularization term of the objective function for now (adding it to the following development is trivial), and let us call $L(\mathbf{W}, \mathbf{H},\R)$ the objective function without regularization. 
We want to be able to update the features of one user or of one item in a time independent of the size of the dataset ($n,m,|\R|$).
In the remainder, we  show that it is possible to compute $\partial L / \partial \wi$ and $\partial L / \partial \hj$ with a complexity linear in the number of ratings provided by user $i$ ($|\Ri|$) or given to item $j$ ($|\Rj|$), respectively.
On most datasets, and for most users and items, we have $|\Ri| \ll m$ and $|\Rj| \ll n$, and, therefore computing the gradient for one user or one item is  fast.

First, let us separate $L$ in $n$ blocks $l_{i}^{w}$ that contain only the terms of $L$ depending on $\wi$:

\begin{align}
l_{i}^{w} = \sum_{j|r_{ij} \in \R} \left( r_{ij} - \wihj \right) ^{2} + \alpha \sum_{j|r_{ij} \notin \R} \left( \wihj \right) ^{2}
\label{eq:l_w_i}
\end{align}

Notice that we have:
\begin{align*}
L = \sum_{i} l_{i}^{w} \quad \text{and} \quad	\frac{\partial L}{\partial \wi} = \frac{\partial l_{i}^{w}}{\partial \wi}
\end{align*}

If we adopt a naive computation,  the second term of Equation (\ref{eq:l_w_i}) is more time expensive because most items are not rated by the user.
However, the sum on unknown ratings (i.e.  $\sum_{j|r_{ij} \notin \R}$), can be formulated as the difference between the sum on all items (i.e. $\sum_{j = 1}^{m}$) and the sum on  rated items only (i.e. $\sum_{j|r_{ij} \in \R}$) . By so doing, the sum on unknown ratings disappears from the computations:

\begin{align}
\sum_{j|r_{ij} \notin \R} \left( \wihj \right) ^{2} &= \sum_{j = 1}^{m} \left( \wihj \right) ^{2} - \sum_{j|r_{ij} \in \R} \left( \wihj \right) ^{2} \\
&= \wi \mathbf{S^h} \wit - \sum_{j|r_{ij} \in \R} \left( \wihj \right) ^{2}
\label{eq:sl_trick}
\end{align}

where we have posed $\mathbf{S^h} = \sum_{j} \hjt\hj$, a $k \times k$ matrix  independent of $i$ (i.e. it is the same matrix for all $l_{i}^{w}$). Assuming that $\mathbf{S^h}$ is known, we can now compute $l_{i}^{w}$ and $\partial L/\partial \wi$ with a complexity of $O(|\Ri|k + k^2)$.
From Equations \ref{eq:l_w_i} and \ref{eq:sl_trick}, we obtain:

\begin{align}
\begin{split}
l_{i}^{w} = \sum_{j|r_{ij} \in \R} \left[ ( r_{ij} - \wihj ) ^{2} - \alpha ( \wihj ) ^{2} \right] + \alpha \wi \mathbf{S^h} \wit 
\end{split}
\label{eq:block}
\end{align}

We can easily derive:

\begin{align}
\begin{split}
\frac{\partial L}{\partial \wi} = -2 \sum_{j|r_{ij} \in \R} \left[ r_{ij} - (1 - \alpha) \wihj \right] \hj + 2 \alpha \wi \mathbf{S^h}
\end{split}
\label{eq:gradient}
\end{align}

Symmetrically, if $\mathbf{S^w} = \sum_{i} \wit\wi$, we have:

\begin{align}
\begin{split}
l_{j}^{h} = \sum_{i|r_{ij} \in \R} \left[ ( r_{ij} - \wihj ) ^{2} - \alpha ( \wihj ) ^{2} \right] + \alpha \hj \mathbf{S^w} \hjt 
\end{split}
\label{eq:fast_block}
\end{align}
and:
\begin{align}
\begin{split}
\frac{\partial L}{\partial \hj} = -2 \sum_{i|r_{ij} \in \R} \left[ r_{ij} - (1 - \alpha) \wihj \right] \wi + 2 \alpha \hj \mathbf{S^w}
\end{split}
\label{eq:fast_gradient}
\end{align}

Assuming that $\mathbf{S^w}$ is known, the complexity of computing $l_{j}^{h}$ or $\partial L/\partial \hj$ is now $O(|\Rj|k + k^2)$, and the complexity of computing it for every $j \in \{1, \ldots, m\}$ is $O(|\R|k + k^2)$.

\subsection{Absolute Loss}\label{sec:absolute-loss}

A similar development can be done when the squared loss is replaced by the absolute loss.
With the absolute loss, $L$ becomes:

\begin{align*}
\begin{split}
L = \sum_{i,j|r_{ij} \in \R} \left| r_{ij} - \wihj \right| + \alpha \sum_{i,j|r_{ij} \notin \R} \left| \wihj \right|
\end{split}
\end{align*}

As with the \SL, we divide $L$ into $l_{i}^{w}$ and $l_{j}^{h}$. 

\begin{align*}
l_{i}^{w} = \sum_{j|r_{ij} \in \R} \left| r_{ij} - \wihj \right| + \alpha \sum_{j|r_{ij} \notin \R} \left| \wihj \right|
\end{align*}

As with the \SL, we will change the expression of $l_{i}^{w}$ to remove the sum over all unknown ratings, but in this case we have to impose non-negativity of the features to go further.
If $\mathbf{W}, \mathbf{H} \geq 0$, we have $\left| \wihj \right| = \wihj$, and therefore:

\begin{align}
\sum_{j|r_{ij} \notin \R} \left| \wihj \right| &= \sum_{j = 1}^{m} \wihj - \sum_{j|r_{ij} \in \R} \wihj \\
&= \wi \left( \sum_{j = 1}^{m} \hjt \right) - \sum_{j|r_{ij} \in \R} \wihj
\label{eq:al_trick}
\end{align}

Here, instead of $\mathbf{S^w}$ and $\mathbf{S^h}$, we will define $\mathbf{s}_{w} = \sum_{i = 1}^{n} \wi$ and $\mathbf{s}_{h} = \sum_{j = 1}^{m} \hj$.
We can now express $l_{i}^{w}$ and $\partial L/ \partial \wi$ efficiently:

\begin{align}
l_{i}^{w} = \sum_{j|r_{ij} \in \R} \left( \left| r_{ij} - \wihj \right| - \alpha \wihj \right) + \alpha \wi \mathbf{s}_{h}^{T} 
\label{eq:al_fast_block}
\end{align}
so that:
\begin{align}
\frac{\partial L}{\partial \wi} = \sum_{j|r_{ij} \in \R} \left( \text{sign}\left(\wihj - r_{ij} \right) - \alpha \right) \hj + \alpha \mathbf{s}_{h}^{T}
\label{eq:al_fast_gradient}
\end{align}
where $\text{sign}(x) = x/|x|$ if $x \neq 0$, and equals $0$ otherwise.
Assuming $\mathbf{s}_{h}$ is known, the complexity of computing $l_{i}^{w}$ or $\partial L/ \partial \wi$ is now $O(|\Ri|k)$.
The corresponding expression of $l_{j}^{h}$ and $\partial L/ \partial \hj$ is trivial, and the complexity to compute them is $O(|\Rj|k)$.

\subsection{Generalized Kullback-Leibler Divergence}\label{sec:kullback-leibler}

For the sake of demonstration on other common loss functions in matrix factorization, we show here the applicability of the sparsity trick on the \GKL~(GKL)~\cite{mult_NMF, lin2007projected}.
We do not elaborate further on this function in the rest of the paper.

The \GKL\ is defined as follows:

\begin{align}
D(r_{ij}||\wihj) = r_{ij} \log (\frac{r_{ij}}{\wihj}) - r_{ij} + \wihj
\label{eq:gkl}
\end{align}

The GKL is not defined when $r_{ij} = 0$. In the following we  extend the GKL by using its limit value:

\begin{align}
D(0||\wihj) := \lim\limits_{r \to 0} D(r||\wihj) = \wihj
\label{eq:gkl_0}
\end{align}

Using Equation \ref{eq:gkl} and \ref{eq:gkl_0}, $L$ becomes:

\begin{align*}
\begin{split}
L = \sum_{i,j|r_{ij} \in \R} \left( r_{ij} \log (\frac{r_{ij}}{\wihj}) - r_{ij} + \wihj \right) + \alpha \sum_{i,j|r_{ij} \notin \R} \wihj
\end{split}
\end{align*}

We now follow the same development as with the other losses.
We define $l_{i}^{w}$:

\begin{align*}
\begin{split}
l_{i}^{w} = \sum_{j|r_{ij} \in \R} \left( r_{ij} \log (\frac{r_{ij}}{\wihj}) - r_{ij} + \wihj \right) + \alpha \sum_{j|r_{ij} \notin \R} \wihj
\end{split}
\end{align*}

In the case of the GKL, the process to remove the sum on unknown ratings is the same as with the \AL, except that in absence of absolute value we do not have to impose non-negativity of the features:

\begin{align}
\sum_{j|r_{ij} \notin \R} \wihj = \wi \mathbf{s}_{h}^{T} - \sum_{j|r_{ij} \in \R} \wihj
\label{eq:gkl_trick}
\end{align}

This leads to:

\begin{align}
\begin{split}
l_{i}^{w} = & \sum_{j|r_{ij} \in \R} \left(r_{ij} \log (\frac{r_{ij}}{\wihj}) - r_{ij} + (1 - \alpha) \wihj \right) \\ 
& + \alpha \wi \mathbf{s}_{h}^{T} 
\end{split}
\end{align}

Now we can easily derive $\partial L/\partial \wi$:

\begin{align}
\frac{\partial L}{\partial \wi} = \sum_{j|r_{ij} \in \R} \left( - \frac{r_{ij}}{\wihj}  + (1 - \alpha) \right) \hj + \alpha \mathbf{s}_{h}^{T}
\end{align}

The corresponding expression of $l_{j}^{h}$ and $\partial L/ \partial \hj$ is obtained symmetrically.
As with the \AL, the complexity of computing $l_{i}^{w}$ or $\partial L/ \partial \wi$ is now $O(|\Ri|k)$ (it is $O(|\Rj|k)$ for $l_{j}^{h}$ and $\partial L/ \partial \hj$).

\subsection{Static and Dynamic Factorization}\label{sec:static-dynamic}

We introduce an online algorithm to learn the latent factors  from the input data in a static setting, and show how  it can accommodate updates in a dynamic setting.

\subsubsection{Static Factorization}
In order to factorize a whole new set of data we propose to use a randomized block coordinate descent \cite{richtarik2014iteration}.
At each iteration, all the users and items are traversed in a random order.
For each of them a gradient step is performed on their features while keeping the other features constant.

We can use a line search \cite{boyd2009convex} to determine the size of the gradient step because the variation of $L$ for a modification of $\wi$ is entirely determined by $l_{i}^{w}$ and can therefore be computed efficiently.
Line search allows to avoid the burden of tuning the step size, proper to stochastic gradient descent (SGD) methods \cite{duchi2011adaptive}.
Moreover, using line search guarantees the convergence of the value of the objective function.
Indeed, each gradient step decreases (or rather \emph{cannot increase}) the objective function which is bounded from below.
This implies that the variation of the objective function converges to zero.

The complete procedure for the factorization through randomized block coordinate descent is summarized in Algorithm \ref{alg:static}.

\spara{Complexity.}
In the case of the \SL, the computation of fast gradient step relies on knowing $\mathbf{S^w}$ and $\mathbf{S^h}$.
Their initial value is computed in $O(nk^2)$ and $O(mk^2)$, respectively, and the cost of updating them after each gradient step is $O(k^2)$.
The total complexity of an iteration of our algorithm is therefore $O(|\R|k + (n + m)k^2)$, as good as the best factorization methods that do not use priors on unknown ratings \cite{gemulla2011large}.

In the case of the \AL~and \GKL, the computation uses $\mathbf{s}_{w}$ and $\mathbf{s}_{h}$.
Their initial value is computed in $O(nk)$ and $O(mk)$, while the cost of updating them is $O(k)$.
The total complexity of one iteration then becomes $O(|\R|k + (n + m)k)$, which is lower than the squared loss' complexity.
However, this usually comes at a cost on the performance of the results, as we will show in the experiments in Section~\ref{sec:experiments}.

\begin{algorithm}[!t]
\caption{Randomized block coordinate descent}
\label{alg:static}
\begin{algorithmic}[1]
\REQUIRE $\,$ \\
 -- The ratings $\R$.\\
 -- The number of features $k$.\\ \algrule
\STATE Initialize $\mathbf{W}$ and $\mathbf{H}$.
\STATE Compute $\mathbf{S^w}$ and $\mathbf{S^h}$ ($\mathbf{s}_{w}$ and $\mathbf{s}_{h}$)
\WHILE{not converged}
	\FORALL{user and item, traversed in a random order} 
		\STATE In the case of a user ($i$) \textbf{do}
		\STATE \hspace{\algorithmicindent} Perform a gradient step on $\wi$ using line search
		\STATE \hspace{\algorithmicindent} Update $\mathbf{S^w}$ ($\mathbf{s}_{w}$)
		\STATE In the case of an item ($j$) \textbf{do}
		\STATE \hspace{\algorithmicindent} Perform a gradient step on $\hj$ using line search
		\STATE \hspace{\algorithmicindent} Update $\mathbf{S^h}$ ($\mathbf{s}_{h}$)
	\ENDFOR
\ENDWHILE
\end{algorithmic}
\end{algorithm}

\subsubsection{Fast Updates}
The expressions of $l_{i}^{w}$, $l_{j}^{h}$, and their gradients (Equations (\ref{eq:block}), (\ref{eq:gradient}), (\ref{eq:fast_block}) and (\ref{eq:fast_gradient})) allow us to compute the latent representations of one user or one item in a time independent of the number of users and items in the system.
We can use that ability to design a simple algorithm for updating an existing factorization when a new rating is added to $\R$:
If user $i$ rates item $j$, we iteratively perform gradient steps for $\wi$ and $\hj$, keeping all other features constant.
This relies on the assumption that a new rating will only affect significantly the user and item that are directly concerned with it.
Although this assumption can be disputed, we will show in our experiments (Section \ref{sec:dynamic}) that our update algorithm produces recommendations of stable quality, indicating that limiting our updates to the directly affected users and items does not degrade the factorization over time.

When ratings are produced by new users or given to new items, a new set of features for that user or item is created before performing the local optimization.
Various initialization strategies could be explored here.
However, as we show in our experimental results, assigning a random value to one of the features and setting the others to zero performs well in practice.
The update procedure is summarized in Algorithm \ref{alg:update}.

\spara{Complexity.}
As mentioned earlier, our update algorithm is independent of the number of users or items in the system, making it suitable for very large datasets.
Each iteration of the update algorithm is composed of two gradient steps (one on the user's features, and one on the item's features).
In particular, the complexity of one iteration is $O((|\Ri| + |\Rj|)k + k^2)$ for the \SL, and only $O((|\Ri| + |\Rj|)k)$ for the \AL~and the GKL.
This difference in complexity becomes significant when $k$ is large with regards to the average number of ratings per user and per item.

Updates based on classic SGD methods have an even smaller complexity ($O(k))$), but we will show in Section \ref{sec:experiments} that our method produces recommendations of much higher quality, while still being able to satisfy applications requiring low-latency updates.

\begin{algorithm}[!t]
\caption{Update algorithm}
\label{alg:update}
\begin{algorithmic}[1]
\REQUIRE $\,$ \\
 -- The new rating $r_{ij}$.\\
 -- The ratings of user $i$ ($\Ri$) and of item $j$ ($\Rj$).\\ \algrule
\STATE If $\wi$ ($\hj$) does not exist, initialize it (for example by setting a random feature to 1).
\STATE Add $r_{ij}$ to $\Ri$ and $\Rj$.
\WHILE{not converged}
\STATE{Perform a gradient step on $\wi$ using line search}
\STATE{Update $\mathbf{S^w}$ ($\mathbf{s}_{w}$)}
\STATE{Perform a gradient step on $\hj$ using line search}
\STATE{Update $\mathbf{S^h}$ ($\mathbf{s}_{h}$)}
\ENDWHILE
\end{algorithmic}
\end{algorithm}
\section{Experiments}
\label{sec:experiments}

We perform several experiments to demonstrate the following key points:
\begin{squishlist}
\item Using priors on the unknown values leads to overall improved quality of ranking, in a static or dynamic setting.
\item The quality does not degrade with time, i.e., as more updates are added, the model does not lose accuracy.
\item Our methods can outperform traditional techniques on various large datasets.
\end{squishlist}

In our experiments, we test the performance of the \SL~(SL) and the \AL~(AL) with and without prior on unknown values.
In Section~\ref{sec:exp-setup} we describe our experimental setup: the benchmarked datasets used, the performance metrics recorded and how we tune the various parameters of the models tested during the experiments.
Then, in Sections~\ref{sec:static} and~\ref{sec:dynamic} we describe the results of our methods in a \emph{static} and \emph{dynamic} learning setting, respectively, and how they compare with state-of-the-art methods.
In Section~\ref{sec:delay} we illustrate the importance of fast updates by studying the impact of having a delay between the arrival of new ratings and the update of the factorization.
In Section~\ref{sec:exp-param-influence} we investigate in depth the influence of parameter values selected in the two loss functions (squared and absolute loss).
Finally,  details allowing the reproducibility of the results are given in Section~\ref{sec:exp-reproducibility}.

\subsection{Experimental Setup}\label{sec:exp-setup}

Here we briefly describe the experimental setup used for the static and dynamic learning and how the parameters of the different methods are tuned.

\subsubsection{Datasets}

During the experiments, we use three datasets with distinct features.
Table~\ref{table:datasets} summarizes the characteristics of these datasets which provide different challenges to the recommendation task:

\begin{squishlist}
\item \dataset{Movielens}: This is the well-known movie ratings dataset produced by the Grouplens project.
We use the version containing 1 million ratings, with at least 20 ratings for each user.
\item \dataset{FineFoods}: This is a collection of ratings about food products extracted from the Amazon comments~\cite{mcauley2013hidden}.
The dataset is much sparser than \dataset{Movielens}, with most users having only a handful of ratings, making it a very hard dataset for the recommendation task.
\item \dataset{AmazonMovies}: This is a larger collection of ratings extracted from the movie section of Amazon~\cite{mcauley2013hidden}.
This dataset is also sparser than \dataset{Movielens}, although not as sparse as \dataset{FineFoods}.
\end{squishlist}

\begin{table}[t]
\caption{Characteristics of the datasets used.}
\centering
\normalsize
\tabcolsep=0.4cm
\begin{tabular}{lrrr}
\toprule
Dataset			&	Users	&	Items	&	Ratings	\\
\midrule
\dataset{Movielens}	&	6,040	&	3,706	&	1,000,209	\\
\dataset{FineFoods}	&	256,059	&	74,258	&	568,454	\\
\dataset{AmazonMovies}	&	889,176	&	253,059	&	7,831,442	\\
\bottomrule
\end{tabular}
\label{table:datasets}
\end{table}

\subsubsection{Evaluation Metrics}

We measure two standards metrics used in ranking evaluation:  (1) Normalized Discounted Cumulative Gain (NDCG) \cite{balakrishnan2012collaborative, lee2014local} and, (2) area under ROC curve (AUC) \cite{ALS_unknown_weighting, rendle2009bpr, transductive_NMF}.

NDCG rewards methods that rank items with the highest observed rating at the top of the ranking.
The \emph{discounted} aspect of NDCG comes from the fact that relevant items ranked at low positions of the ranking contribute less to the final score than relevant items at top positions.

In the static experiments, we also report  the NDCG computed on the rated items only. 
This metric does not consider the real world case scenario which consists of ranking all items since we do not know in advance which item will be rated or not. 
Intuitively, by biasing our objective through the introduction of priors on unknown rating we may loose performance when ranking rated items only, while performing better when considering all the items.  

We use AUC to evaluate the ability of the different methods to predict which items are going to be rated.
AUC measures whether the items whose ratings were held out during learning are ranked higher than unrated items.
The perfect ranking has an AUC of 1, while the average AUC for random ranking is $0.5$.

\subsubsection{Parameter Tuning}

Table~\ref{table:parameters} shows the parameters of the various models and the values tested during parameter tuning.
For each test, the parameters' values producing the best ranking on the validation sets (measured by NDCG for the static test and AUC for the dynamic test) were selected to be used.
See Sections \ref{sec:static} and \ref{sec:dynamic} for the description of the validation sets.

\begin{table}[t]
\caption{List of the parameters of each method, and set of values tested during the parameters tuning of the squared loss (SL) and absolute loss (AL), with and without prior on unknown values, as well as the multiplicative update algorithm (Mult-NMF), Alternating Least Square (ALS-UV), and Vowpal Wabbit (VW).
$k$: number of features, $\lambda$: regularization coefficient, $\rho$: unknown/known influence ratio, $\gamma$: learning rate.}
\centering
\tabcolsep=0.3cm
\begin{tabular}{lcl}
\toprule
Method					&	Parameter		&	Tested Values	\\
\midrule
\multirow{3}*{\parbox{1.4cm}{SL/AL with prior}} & $k$	& 5, 10, 20, 50, 100, 200 \\
						&	$\lambda$		& 0, 0.01, 0.1, 1, 10 \\
						&	$\rho$		& 0.3, 0.7, 1, 2 \\
\midrule
\multirow{2}*{\parbox{1.9cm}{SL/AL without prior}}	& $k$ & 5, 10, 20, 50, 100, 200 \\
						&	$\lambda$		& 0, 0.01, 0.1, 1, 10 \\
\midrule
\multirow{2}*{ALS-UV}			&	$k$			& 20, 50, 100, 200, 500 \\
						&	$\lambda$		& 0, 0.001, 0.01, 0.05, 0.1 \\
\midrule
Mult-NMF						&	$k$			& 20, 50, 100, 200, 500 \\
\midrule
\multirow{3}*{\parbox{1.5cm}{\VW}}	&	$k$ 	& 20, 50, 100, 200, 500 \\
						&	$\lambda$		& 0 1e-5 1e-2 \\
						&	$\gamma$ 	& 0.01, 0.02, 0.05, 0.1, 0.2 \\
\bottomrule
\end{tabular}
\label{table:parameters}
\end{table}

\subsection{Static Learning}\label{sec:static}

\spara{Research question.}
In a static mode, we test to which extent  using a prior on unknown ratings improves the ranking of items when  recommended  to users. 

\spara{Process followed.}
The test set was constructed by randomly selecting 1000 users, and splitting the ratings of those users in half, keeping the first 50\% of the ratings in the training set, according to timestamp, and the last 50\% in the test set.
The same process (selecting 1000 users and splitting their ratings) was then applied three times on the training set in order to create three training/validation pairs of sets.
On each run, the parameters producing on average the best NDCG over the three validation sets were then used to factorize the full training set, and evaluated on the test set. 

\spara{Baseline.}
We report the results achieved by two traditional well-known algorithms: UV matrix decomposition solved with Alternating Least Square (ALS-UV) \cite{zhou2008large}, and non-negative matrix factorization with the multiplicative update algorithm (Mult-NMF) \cite{mult_NMF}.
Both ALS-UV and Mult-NMF use the \SL.

\spara{Results.}
The results, averaged over 10 runs, are shown in Table \ref{table:static_results}. 
We can observe that for both the \SL~and the absolute loss, and on all datasets, by adding a  prior on the unknown ratings we improve significantly the rankings of the items recommended to users over rankings obtained by the same techniques when they do not put a prior on the unknown ratings (and also over rankings obtained by state-of-the-art approaches ALS-UV and Mult-NMF).
In particular, our implementation of the \SL~with prior outperforms all other methods, as confirmed by a Mann-Withney U test with a confidence level of 1\%.

On the \dataset{Movielens} dataset, the results of Mult-NMF and ALS-UV are, as expected, similar to the ones of our implementation of the \SL~without prior.
Indeed, those methods optimize the same objective function, and differ only by their algorithm.
Interestingly, on the sparser \dataset{FineFoods} and \dataset{AmazonMovies} dataset, our randomized block coordinate gradient descend method outperforms Mult-NMF and ALS-UV, even without prior on the unknown ratings.

Furthermore, in the three tested data sets, when only considering the rated items, the loss in ranking performance is never significant (see NDCG-RI with and without prior in Table \ref{table:static_results}).
In other words, while improving on the global ranking, the performance does not deteriorate when considering only the subset of rated items.

\begin{table} 
\centering
\scriptsize
\caption{Comparison of our introduced algorithm in static learning on the datasets \dataset{Movielens}, \dataset{FineFoods} and \dataset{AmazonMovies}. 
Values in bold hold for the method that outperform all the other methods according to a Mann-Withney U test with a confidence level of 1\%.
Average values are shown alongside their standard deviation over 10 runs.}
\tabcolsep=0.07cm
\begin{tabular}{l c c c}
\hline
			& \multicolumn{3}{c}{\dataset{Movielens}}						\\
			& NDCG-RI	 & NDCG 					& AUC					\\
\hline
SL w/ prior 	&0.885 $\pm$ 0.0014& \bf{0.5046} $\pm$ 0.0013	& \bf{0.8695} $\pm$ 0.0012		\\
SL w/o prior & 0.886  $\pm$ 0.0015 &  0.3597 $\pm$ 0.0012		& 0.6548 $\pm$ 0.0014		\\
AL w/ prior	& 0.8683 $\pm$ 0.0030 & 0.4452 $\pm$ 0.0009		& 0.8134 $\pm$ 0.0011			\\
AL w/o prior & 0.8794  $\pm$  0.0031 &0.3801 $\pm$ 0.0106		& 0.6927 $\pm$ 0.0322	\\
Mult-NMF		& 0.8433 $\pm$ 0.0007 & 0.3758 $\pm$ 0.0006		& 0.7011 $\pm$ 0.0009		\\
ALS-UV		& 0.8332 $\pm$ 0.0014 & 0.3292 $\pm$ 0.0004		& 0.5839 $\pm$ 0.0005		\\
\hline
			& \multicolumn{3}{c}{\dataset{FineFoods}}						\\
			& NDCG-RI	 & NDCG 					& AUC					\\
\hline
SL w/ prior	&0.887 $\pm$ 0.0016&  \bf{0.1237} $\pm$ 0.0039	& \bf{0.8452} $\pm$ 0.0074	\\
SL w/o prior	& 0.888  $\pm$ 	0.0158 & 0.1023 $\pm$ 0.0022		& 0.8314 $\pm$ 0.0058		\\
AL w/ prior	&0.8722  $\pm$  0.0142 & 0.1026 $\pm$ 0.0030		& 0.8412 $\pm$ 0.0047		\\
AL w/o prior &0.8730  $\pm$ 0.0260	&  0.0923 $\pm$ 0.0008		& 0.7294 $\pm$ 0.0143		\\
Mult-NMF		& 0.8476 $\pm$  0.0084	& 0.0830 $\pm$ 0.0008		& 0.3403 $\pm$ 0.0052		\\
ALS-UV		& 0.8653 $\pm$  0.025	& 0.0873 $\pm$ 0.0009		& 0.5485 $\pm$ 0.0114		\\
\hline
			& \multicolumn{3}{c}{\dataset{AmazonMovies}}					\\
			& NDCG-RI	 & NDCG 					& AUC					\\
\hline
SL w/ prior	&0.8992  $\pm$ 0.0101 &  \bf{0.1887} $\pm$ 0.0088	& \bf{0.9276} $\pm$ 0.0031	\\
SL w/o prior & 0.9035  $\pm$ 0.0089 & 0.1103 $\pm$ 0.0008		& 0.8656 $\pm$ 0.0033				\\
AL w/ prior	&0.8804 $\pm$ 0.0077 &  0.1348 $\pm$ 0.0035		& 0.8634 $\pm$ 0.0045		\\
AL w/o prior	&0.8854  $\pm$  0.0102 & 0.1002 $\pm$ 0.0012		& 0.7625 $\pm$ 0.0051			\\
Mult-NMF		&  0.8498 $\pm$   0.0026 & 0.0959 $\pm$ 0.0004		& 0.6330 $\pm$ 0.0040		\\
ALS-UV		&  0.8658 $\pm$   0.0034 & 0.0906 $\pm$ 0.0003		& 0.6601 $\pm$ 0.0061		\\
\hline
\end{tabular}
\label{table:static_results}
\end{table}

\subsection{Dynamic Learning}\label{sec:dynamic}

\spara{Research question.}
In this section, we target two research questions:

\begin{enumerate}
\item We test whether our update algorithm is able to sustain stable quality of recommendations over time;
\item We test to which extent using priors on unknown ratings improves the ranking of recommended items when the model is updated each time a new instance is encountered. By so doing, the system is evaluated on more realistic scenarios where the cases of cold items and users are considered as well.
\end{enumerate}

\spara{Process followed.}
We order the ratings by timestamps and separate the ratings in three blocks: first the training, then the validation and finally the test block (see Table \ref{table:datasets_block_size} for the size of each block in the different datasets).

\begin{table}[h]
\caption{Number of ratings in each block for the dynamic learning.}
\centering
\tabcolsep=0.15cm
\begin{tabular}{lrrr}
\toprule
Dataset			&	Training	&	Validation	&	Test	\\
\midrule
\dataset{Movielens}	&	500,000	&	100,000	&	100,000	\\
\dataset{FineFoods}	&	400,000	&	100,000	&	68,000	\\
\dataset{AmazonMovies}	&	5,000,000	&	1,000,000	&	1,000,000	\\
\bottomrule
\end{tabular}
\label{table:datasets_block_size}
\end{table}

The evaluation is performed as follows: an initial model is built based on all the ratings present before the test block, then, for each rating of the test block, two steps are performed in the following order:
\begin{enumerate}
	\item The current model is evaluated by computing the AUC over the new (user, item) pair. 
	Notice that in this case, computing the AUC means computing the proportion of items not yet rated by the user that the model ranks lower than the item that was just rated.
	An AUC of 1 means that the new item was the top recommendation of the method for that user.
	\item The model is updated using the new rating. It is worth noticing that the rating may concern a new user or item, and, therefore, features for that new user/item have to be added to the model.
\end{enumerate}

Parameter tuning is done as described above, but starting at the beginning of the validation block and ending before the test block. 
The values of parameters tested are the same as in the static test (see Table \ref{table:parameters}).

\spara{Baseline.}
We compared our methods to Vowpal Wabbit (\VW).
\VW\ is a machine learning framework solving different optimization problems for classification and ranking, by implementing a carefully optimized, stochastic gradient descent (SGD) using feature hashing \cite{weinberger2009feature} and adaptive gradient steps \cite{duchi2011adaptive}.
We are using the \VW's implementation of low-rank interactions\footnote{\url{https://github.com/JohnLangford/vowpal_wabbit/tree/master/demo/movielens}} based on factorization machines~\cite{DBLP:conf/icdm/Rendle10}.

\begin{figure}
\centering
	\includegraphics[scale=0.5]{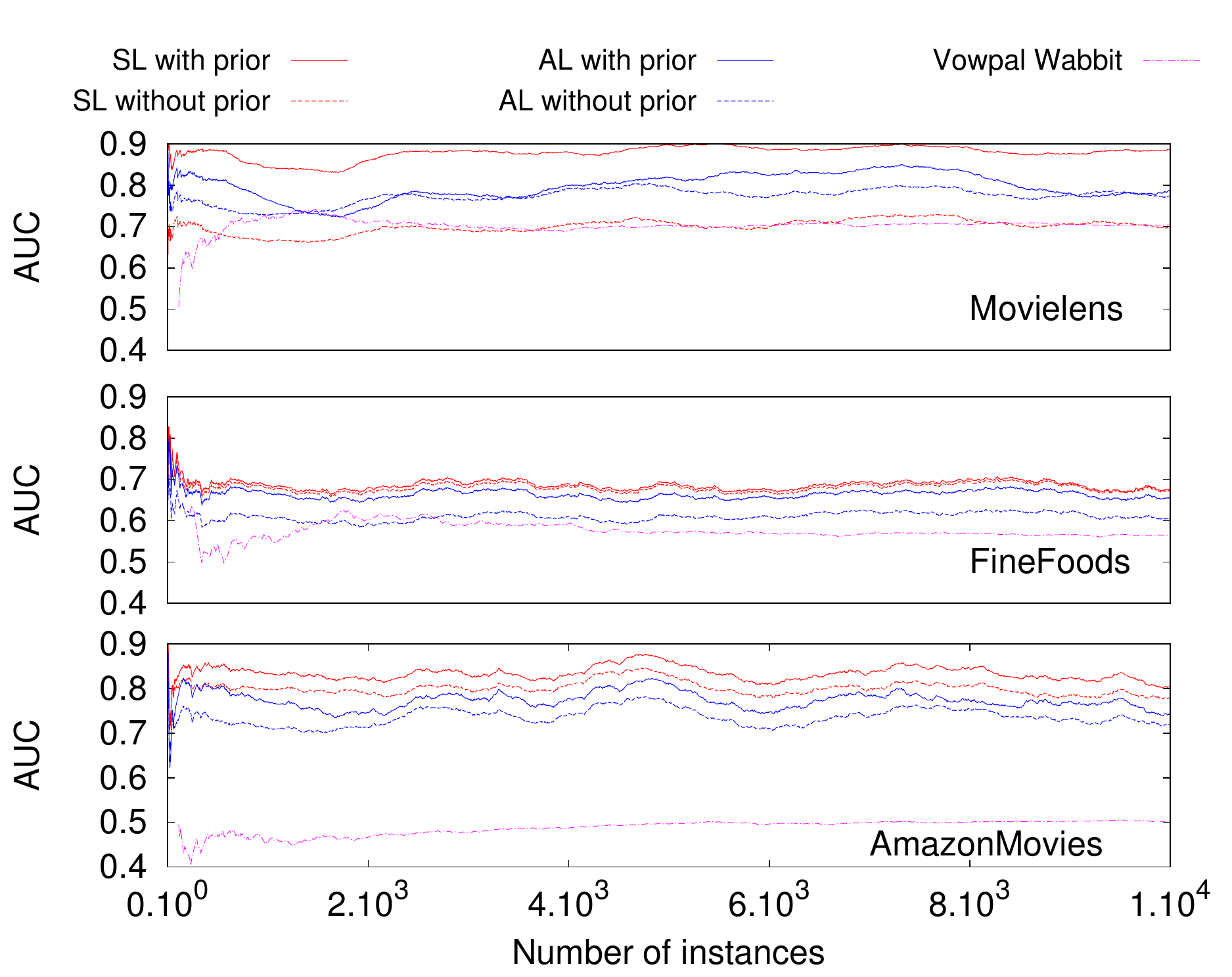}
\caption{Performance comparison with respect to average AUC for the various methods tested in dynamic learning on \dataset{Movielens}, \dataset{FineFoods} and \dataset{AmazonMovies}.
Results are averaged over 20 runs.}
\label{fig:dynamic_results-movielens}
\end{figure}

\spara{Results.}
Figure~\ref{fig:dynamic_results-movielens} shows how the average AUC evolves as new ratings enter the system.
We first observe that the quality of the results does not decrease over time, indicating that our update algorithm can work for long periods of time without propagating or amplifying errors.
As in the static experiment, we confirm that adding a  prior on unknown ratings improves the quality of the ranking and, again, this is maintained across time.
Moreover, the SGD approach of \VW\ is outranked in each dataset by our approach with prior.

\subsection{Impact of delayed updates}\label{sec:delay}

\spara{Research Question.}
We test the performance of delayed models produced by our methods in delivering recommendations to users.

\spara{Process followed.}
In order to address this question, we simulate a recommender system that is not able to incorporate new ratings in the model as soon as they enter the system.
To do so, we modify the process of dynamic learning presented in Section~\ref{sec:dynamic} to impose a delay between the arrival of a new rating and the update of the factorization.
More precisely, after the $i^{\text{th}}$ rating is given by a user, the model is updated up to the $(i - d)^{\text{th}}$ rating ($d$ being the arbitrary delay).
This way, the model is always $d$ ratings behind the last one arrived (the ratings are sorted by real time of arrival).
In real applications, the delay would probably vary, depending on the level of activity of the users.
However, this experiment gives a first impression of the impact of delays on the recommendation task.

\begin{figure}
\centering
	\includegraphics[scale=0.7]{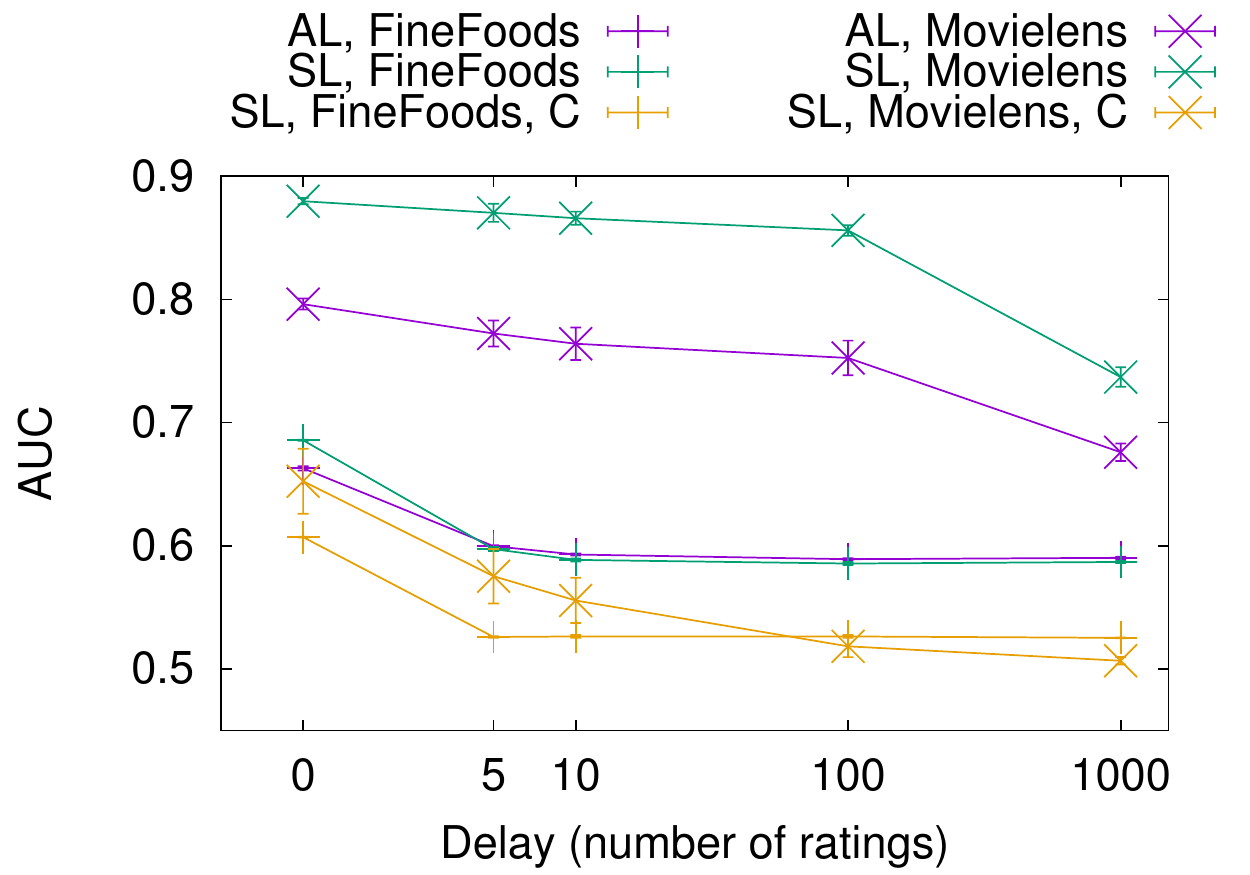}
\caption{Average AUC of the \SL~and \AL~with prior on \dataset{Movielens} and \dataset{FineFoods} for various delays $d$, imposed as a number of ratings that the model is behind the current rating.}
\label{fig:delay}
\end{figure}

\spara{Results.}
Figure~\ref{fig:delay} shows the impact of a delay on the average AUC of the \SL~and \AL~with prior for a dense dataset like \dataset{Movielens} and a sparse dataset like \dataset{FineFoods}.
We observe that even a small delay can affect the quality of the recommendation, depending on the characteristics of the data.
For \dataset{Movielens}, if the model is behind by $5-10$ ratings, the average AUC drops by $3\%$, and it goes down by about $14\%$ when the model is behind by $1000$ ratings, and this applies to both loss functions.
On the other hand, for the much sparser \dataset{FineFoods}, the effect is more apparent.
With only $5-10$ ratings behind, the model's AUC already drops by $10\%$.

To show the effect of fast updates on weakly-engaged (or cold) users, we also report the impact of delays on those users for both \dataset{Movielens} and \dataset{FineFoods} with the \SL\ which performs best (Figure~\ref{fig:delay}, Cold Users). 
We define such users as the ones that rated at most two items.
As hypothesized, the cost induced by delayed predictions (for five ratings delayed) is higher for cold users.
We observe a relative drop in AUC of 11.8\% and 13.4\% for weakly-engaged users on \dataset{Movielens} and \dataset{FineFoods}, respectively, while when considering all the users, the relative drop is 1.1\% and 12.9\%, respectively.

In such sparse scenarios, cold users perform only a handful of actions before deciding to abandon the site or not.
Therefore, it is important to consider cold users in the model as soon as they arrive, to keep them engaged by fast, efficient and good recommendations.

\subsection{Parameter Analysis}\label{sec:exp-param-influence}

\spara{Research Question.}
We test to which extent the number of features ($k$), the weight of the prior ($\rho$) and the regularization coefficient ($\lambda$) affect the AUC and the runtime per update on our loss functions.

\begin{figure*}[htbp]
\begin{center}
	\includegraphics[scale=0.45]{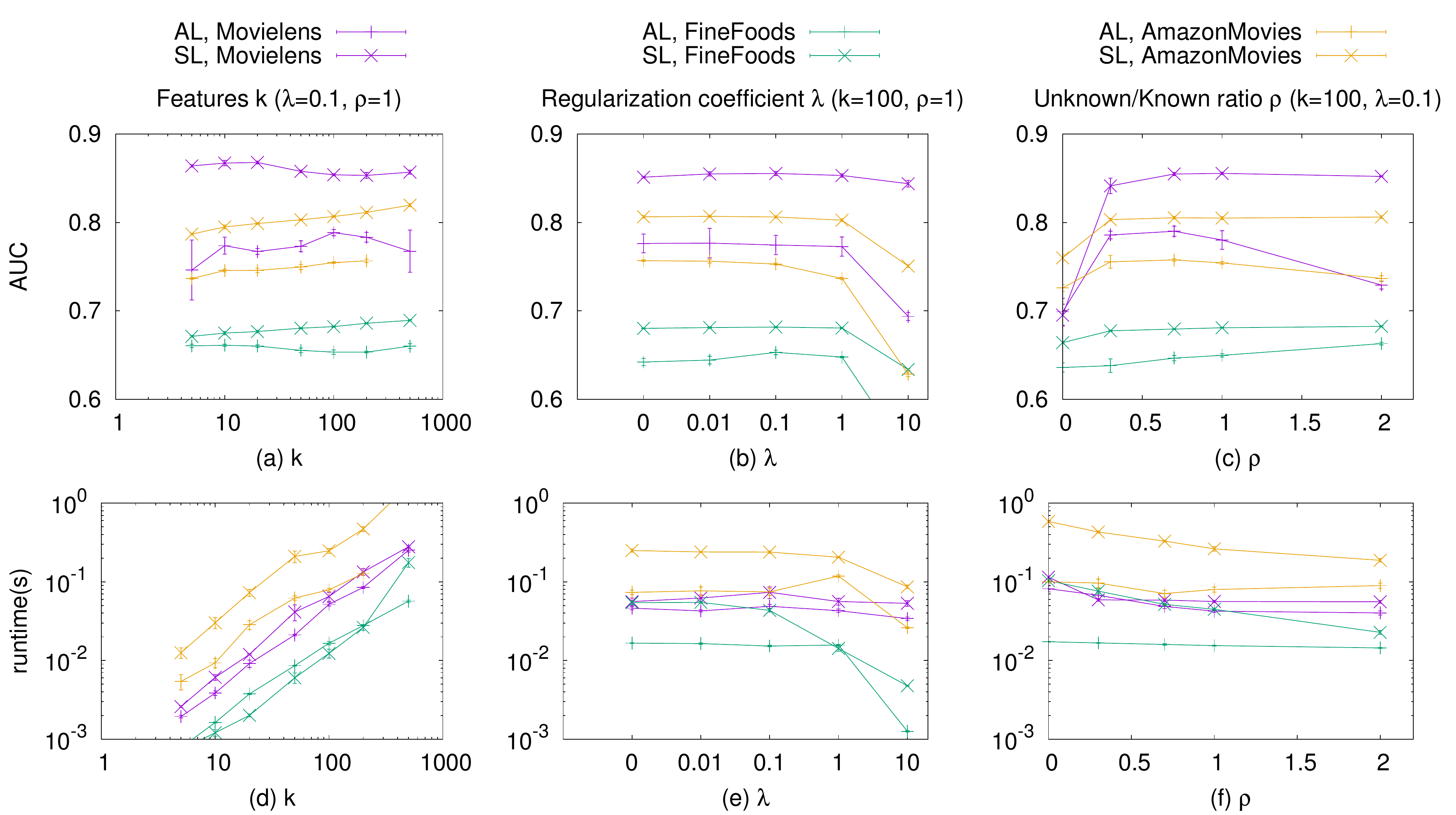}
	\caption{Influence of parameter values on the AUC and runtime for the models produced by the \SL~(SL) and \AL~(AL). Y-error bars declare a standard deviation on the average value of each metric.}
\label{fig:parameter-influence}
\end{center}
\end{figure*}

Figure \ref{fig:parameter-influence} shows the results of this investigation for different values of these parameters, for both \SL~and \AL~and on each dataset.
The results are obtained using the dynamic learning process.

\spara{Number of features.}
Concerning the quality of ranking (AUC), we observe the usual overfitting/under-fitting trade-off (Figure~\ref{fig:parameter-influence}(a)).
The optimal number of features depends on the dataset as well as on the loss function used, suggesting that a careful tuning of that parameter is always needed.

In some cases, speed constraints will force the use a suboptimal number of features.
Indeed, the update runtime heavily depends on the number of features.
Figure~\ref{fig:parameter-influence}(d) suggests a linear relationship between runtime and number of features.
For both losses there is indeed a linear role of the number of features in the theoretical complexity (Section \ref{sec:static-dynamic}).
Notice, however, that the theoretical complexity of the \SL\ also has a quadratic term that becomes dominant for large number of features (with regards to the number of ratings per users).
Also note that while the \SL\ produces better AUC, the \AL\ is able to sustain higher update rates, and can therefore be the loss of choice when speed is the first criterion.

\spara{Regularization coefficient.} 
The influence of $\lambda$ seems rather limited, except for high values that cause both the AUC and the update runtime to drop (Figure\ref{fig:parameter-influence}(b) and (e)).
A small regularization is supposed to increase the quality of the model by reducing overfitting, but this effect is not visible here.
The reason may be that the role of regularization is already taken by the prior on unknown ratings.
Introducing the prior seems to have the side effect of making the regularization obsolete (or redundant).
In fact, we confirm this with the results for $\lambda = 0$ which demonstrate no impact on the quality or runtime.
Again, we see that setting a prior on unknown ratings increases the quality of recommendations without increasing the complexity of the solution.
While it adds a term and a parameter to the objective function, it allows to remove one and its associated parameters.

\spara{Unknown/known influence ratio.}
The ratio $\rho$ influences the performance of the \SL~algorithm in the following way: the AUC increases when a prior on unknown values is added ($\rho > 0$), but the exact value of $\rho$ has little influence (in the observed range) (Figure~\ref{fig:parameter-influence}(c)).
The absolute loss is more sensitive to the value of $\rho$, with the AUC decreasing when $\rho$ becomes too large (on \dataset{Movielens} and \dataset{AmazonMovies}).
However, in both cases, and on all datasets, giving the same weight to the known and unknown ratings ($\rho = 1$) offers a significant improvement over not using a prior, suggesting that $\rho = 1$ can be used as a first guideline, avoiding the burden of further parameter tuning.

The update runtime is also affected by $\rho$, decreasing when $\rho$ increases (Figure~\ref{fig:parameter-influence}(f)).
The explanation can be that the prior on unknown ratings acts as a regularizer, driving features towards 0, and in doing so speeding up the convergence.

\spara{Runtime.}
In general, our technique demonstrates low running time which is heavily dependent on the number of features used, and less on the regularization applied or the ratio of unknown over known values.
These results demonstrate that our method can satisfy applications requiring low-latency updates.

\subsection{Reproducibility of Results}\label{sec:exp-reproducibility}

The implementation of the algorithms introduced in Section~\ref{sec:loss-funtcions} is available on Github:\\
\codeurl.

For both ALS-UV and Mult-NMF we use the implementation of GraphChi, an open source tool for graph computation with impressive performance~\cite{kyrola2012graphchi}.

The code and documentation of Vowpal Wabbit is available on its Github page:\\
\url{https://github.com/JohnLangford/vowpal_wabbit/wiki}.

The \dataset{Amazon} datasets are available on the SNAP webpage:\\
\url{http://snap.stanford.edu/data/index.html}

The \dataset{Movielens} dataset is available on the Grouplens page:\\
\url{http://www.grouplens.org/datasets/movielens/}.
\section{Related Work}\label{sec:related-work}

The problem of recommending products based on the actions and feedback from other users (rather than based on content similarity) is often called \emph{collaborative filtering}, and dates back 20 years ago, with works such as Tapestry~\cite{tapestry} and Grouplens~\cite{grouplens}.
The field is now dominated by methods based on matrix factorization, with algorithms such as ALS~\cite{zhou2008large}, the multiplicative update rule~\cite{mult_NMF}, and the stochastic gradient descent method (SGD)~\cite{gemulla2011large, koren2009matrix}.

The missing at random assumption has yet to get the attention it deserves in collaborative filtering. 
Both ~\cite{missing_at_random} and ~\cite{steck2010training}  have validated the hypothesis  of ratings missing not at random.
Practical propositions for the interpretation of missing data can be found in the fields of one-class collaborative filtering and collaborative filtering based on implicit feedback, where the missing at random assumption is often obviously untenable~\cite{implicit_feedback_confidence, ALS_unknown_weighting, weighting_sampling}.
~\cite{transductive_NMF} offers an interesting approach where missing ratings are considered as optimization variables; they use an EM algorithm to optimize in turn the factorization and the estimation of missing values. Unfortunately, that method has a high complexity, and the proposed approximations that work with large problems remove some of the method's appeal.

None of those works, however, consider the real world, dynamic scenario of continuously observing new ratings, users and items.
Other works \cite{gaillard2015time, rendle2008online} focus on the dynamic update of matrix factorization (mainly through the use of SGD), but those, on the other hand, implicitly rely on the missing at random assumption, and therefore suffer from lower accuracy in predictions. Other state-of-art methods for matrix factorization scale by relying on stochastic gradient computation \cite{chin2015fast, chin2015learning}, while we rely on exact gradient approach. In this work, at the difference of what is mostly seen on scalable machine learning techniques nowadays \cite{dean2012large}, we base our approach on coordinate random block descent to compute exact gradient in order to deal with missing data of large scale matrices.
\section{Conclusions}\label{sec:conclusion}
In this work we proposed a new, simple, and efficient, way to incorporate a prior on unknown ratings in several loss functions commonly used for matrix factorization.
We experimentally demonstrated the importance of adding such a prior to solve the problem of collaborative ranking.
We also tackled the problem of updating the factorization when new users, items and ratings enter the system.
We believe that this problem is central to real applications of recommendation systems, because new users constantly enter those systems and the factorization must be kept up to date to give them recommendations immediately after their first few interactions with the platform.
We offer an update algorithm whose complexity is independent of the size of the data, making it a good approach for large datasets.
In the future, we would like to explore how our methods perform under real workloads of updates with variable arrival rates of ratings per user and item.
Furthermore, we would like to test the performance of our methods in platforms built to analyze streams of data such as Storm, Twitter's Distributed Processing Engines platform.
\section{Acknowledgements}
R. Devooght is supported by the Belgian Fonds pour la Recherche dans l’Industrie et l’Agriculture (FRIA, 1.E041.14).
\bibliographystyle{abbrv}

\begin{thebibliography}{10}

\bibitem{balakrishnan2012collaborative}
S.~Balakrishnan and S.~Chopra.
\newblock Collaborative ranking.
\newblock In {\em Proc. of the 5th ACM WSDM}, pages 143--152, 2012.

\bibitem{boyd2009convex}
S.~Boyd and L.~Vandenberghe.
\newblock {\em Convex optimization}.
\newblock Cambridge university press, 2009.

\bibitem{chin2015fast}
W.-S. Chin, Y.~Zhuang, Y.-C. Juan, and C.-J. Lin.
\newblock A fast parallel stochastic gradient method for matrix factorization
  in shared memory systems.
\newblock {\em ACM Transactions on Intelligent Systems and Technology (TIST)},
  6(1):2, 2015.

\bibitem{chin2015learning}
W.-S. Chin, Y.~Zhuang, Y.-C. Juan, and C.-J. Lin.
\newblock A learning-rate schedule for stochastic gradient methods to matrix
  factorization.
\newblock In {\em Advances in Knowledge Discovery and Data Mining}, pages
  442--455. Springer, 2015.

\bibitem{cremonesi2010performance}
P.~Cremonesi, Y.~Koren, and R.~Turrin.
\newblock Performance of recommender algorithms on top-n recommendation tasks.
\newblock In {\em Proc. of the 4th ACM RecSys}, pages 39--46, 2010.

\bibitem{dean2012large}
J.~Dean, G.~Corrado, R.~Monga, K.~Chen, M.~Devin, M.~Mao, A.~Senior, P.~Tucker,
  K.~Yang, Q.~V. Le, et~al.
\newblock Large scale distributed deep networks.
\newblock In {\em Advances in Neural Information Processing Systems}, pages
  1223--1231, 2012.

\bibitem{duchi2011adaptive}
J.~Duchi, E.~Hazan, and Y.~Singer.
\newblock Adaptive subgradient methods for online learning and stochastic
  optimization.
\newblock {\em The Journal of Machine Learning Research}, 12:2121--2159, 2011.

\bibitem{gaillard2015time}
J.~Gaillard and J.-M. Renders.
\newblock Time-sensitive collaborative filtering through adaptive matrix
  completion.
\newblock In {\em Advances in Information Retrieval}, pages 327--332. Springer,
  2015.

\bibitem{gemulla2011large}
R.~Gemulla, E.~Nijkamp, P.~J. Haas, and Y.~Sismanis.
\newblock Large-scale matrix factorization with distributed stochastic gradient
  descent.
\newblock In {\em Proc. of the 17th ACM SIGKDD}, pages 69--77, 2011.

\bibitem{tapestry}
D.~Goldberg, D.~Nichols, B.~M. Oki, and D.~Terry.
\newblock Using collaborative filtering to weave an information tapestry.
\newblock {\em Communications of the ACM}, 35(12):61--70, 1992.

\bibitem{implicit_feedback_confidence}
Y.~Hu, Y.~Koren, and C.~Volinsky.
\newblock Collaborative filtering for implicit feedback datasets.
\newblock In {\em Proc. of the 8th IEEE ICDM}, pages 263--272, 2008.

\bibitem{koren2009matrix}
Y.~Koren, R.~Bell, and C.~Volinsky.
\newblock Matrix factorization techniques for recommender systems.
\newblock {\em Computer}, 42(8):30--37, 2009.

\bibitem{kyrola2012graphchi}
A.~Kyrola, G.~E. Blelloch, and C.~Guestrin.
\newblock Graphchi: Large-scale graph computation on just a pc.
\newblock In {\em OSDI}, volume~12, pages 31--46, 2012.

\bibitem{mult_NMF}
D.~D. Lee and H.~S. Seung.
\newblock Algorithms for non-negative matrix factorization.
\newblock In {\em Advances in neural information processing systems}, pages
  556--562, 2000.

\bibitem{lee2014local}
J.~Lee, S.~Bengio, S.~Kim, G.~Lebanon, and Y.~Singer.
\newblock Local collaborative ranking.
\newblock In {\em Proc. of the 23rd ACM WWW}, pages 85--96, 2014.

\bibitem{lin2007projected}
C.-J. Lin.
\newblock Projected gradient methods for nonnegative matrix factorization.
\newblock {\em Neural computation}, 19(10):2756--2779, 2007.

\bibitem{little2002statistical}
R.~J. Little and D.~B. Rubin.
\newblock Statistical analysis with missing data.
\newblock 2002.

\bibitem{missing_at_random}
B.~Marlin, R.~S. Zemel, S.~Roweis, and M.~Slaney.
\newblock Collaborative filtering and the missing at random assumption.
\newblock In {\em Proc. of the 23rd Conference on Uncertainty in Artificial
  Intelligence}, 2007.

\bibitem{mcauley2013hidden}
J.~McAuley and J.~Leskovec.
\newblock Hidden factors and hidden topics: understanding rating dimensions
  with review text.
\newblock In {\em Proc. of the 7th ACM RecSys}, pages 165--172, 2013.

\bibitem{ALS_unknown_weighting}
R.~Pan and M.~Scholz.
\newblock Mind the gaps: weighting the unknown in large-scale one-class
  collaborative filtering.
\newblock In {\em Proc. of the 15th ACM SIGKDD}, pages 667--676. ACM, 2009.

\bibitem{weighting_sampling}
R.~Pan, Y.~Zhou, B.~Cao, N.~N. Liu, R.~Lukose, M.~Scholz, and Q.~Yang.
\newblock One-class collaborative filtering.
\newblock In {\em Proc. of the 8th IEEE ICDM}, pages 502--511, 2008.

\bibitem{DBLP:conf/icdm/Rendle10}
S.~Rendle.
\newblock Factorization machines.
\newblock In {\em Proc. of {ICDM} 2010}, pages 995--1000, 2010.

\bibitem{rendle2009bpr}
S.~Rendle, C.~Freudenthaler, Z.~Gantner, and L.~Schmidt-Thieme.
\newblock Bpr: Bayesian personalized ranking from implicit feedback.
\newblock In {\em Proc. of the 25th Conference on Uncertainty in Artificial
  Intelligence}, pages 452--461. AUAI Press, 2009.

\bibitem{rendle2008online}
S.~Rendle and L.~Schmidt-Thieme.
\newblock Online-updating regularized kernel matrix factorization models for
  large-scale recommender systems.
\newblock In {\em Proc. of the 2008 ACM RecSys}, pages 251--258, 2008.

\bibitem{grouplens}
P.~Resnick, N.~Iacovou, M.~Suchak, P.~Bergstrom, and J.~Riedl.
\newblock Grouplens: an open architecture for collaborative filtering of
  netnews.
\newblock In {\em Proc. of the 1994 ACM conference on Computer supported
  cooperative work}, pages 175--186, 1994.

\bibitem{richtarik2014iteration}
P.~Richt{\'a}rik and M.~Tak{\'a}{\v{c}}.
\newblock Iteration complexity of randomized block-coordinate descent methods
  for minimizing a composite function.
\newblock {\em Mathematical Programming}, 144(1-2):1--38, 2014.

\bibitem{transductive_NMF}
V.~Sindhwani, S.~S. Bucak, J.~Hu, and A.~Mojsilovic.
\newblock One-class matrix completion with low-density factorizations.
\newblock In {\em Proc. of the 10th IEEE ICDM}, pages 1055--1060, 2010.

\bibitem{steck2010training}
H.~Steck.
\newblock Training and testing of recommender systems on data missing not at
  random.
\newblock In {\em Proc. of the 16th ACM SIGKDD}, pages 713--722. ACM, 2010.

\bibitem{weinberger2009feature}
K.~Weinberger, A.~Dasgupta, J.~Langford, A.~Smola, and J.~Attenberg.
\newblock Feature hashing for large scale multitask learning.
\newblock In {\em Proc. of the 26th ACM ICML}, pages 1113--1120. ACM, 2009.

\bibitem{zhou2008large}
Y.~Zhou, D.~Wilkinson, R.~Schreiber, and R.~Pan.
\newblock Large-scale parallel collaborative filtering for the netflix prize.
\newblock In {\em Algorithmic Aspects in Information and Management}, pages
  337--348. Springer, 2008.

\end{thebibliography}

\end{document}